\def\year{2021}\relax
\documentclass[letterpaper]{article} 
\usepackage{aaai21}  
\usepackage{times}  
\usepackage{helvet} 
\usepackage{courier}  
\usepackage[hyphens]{url}  
\usepackage{graphicx} 

\usepackage{amsmath,amssymb} 
\usepackage{color}
\usepackage{subfigure}
\usepackage{epstopdf}
\usepackage{algorithm}
\usepackage{algorithmic}
\usepackage{multirow}
\usepackage{array}

\usepackage[switch]{lineno}
\usepackage{natbib}  
\usepackage{caption} 
\title{Online Hashing with Similarity Learning}
\author {
\Large \textbf{Zhenyu Weng, Yuesheng Zhu}\\
Communication and Information Security Laboratory, Shenzhen Graduate School, Peking University\\
wzytumbler@pku.edu.cn, zhuys@pku.edu.cn
}

\begin{document}

\maketitle

\begin{abstract}
Online hashing methods usually learn the hash functions online, aiming to efficiently adapt to the data variations in the streaming environment. However, when the hash functions are updated, the binary codes for the whole database have to be updated to be consistent with the hash functions, resulting in the inefficiency in the online image retrieval process. In this paper, we propose a novel online hashing framework without updating binary codes. In the proposed framework, the hash functions are fixed and a parametric similarity function for the binary codes is learnt online to adapt to the streaming data. Specifically, a parametric similarity function that has a bilinear form is adopted and a metric learning algorithm is proposed to learn the similarity function online based on the characteristics of the hashing methods. The experiments on two multi-label image datasets show that our method is competitive or outperforms the state-of-the-art online hashing methods in terms of both accuracy and efficiency for multi-label image retrieval.
\end{abstract}

\section{Introduction}
The large amount of visual data (e.g. images and videos) that is produced everyday leads to the development of efficient index and retrieval methods. Among these methods, hashing methods that perform approximate nearest neighbor search by mapping high-dimensional data to binary codes have attracted increasing attention~\cite{7360966,7915742}.

\begin{figure*}[!htbp]
\centering

 \includegraphics[width=12.2cm]{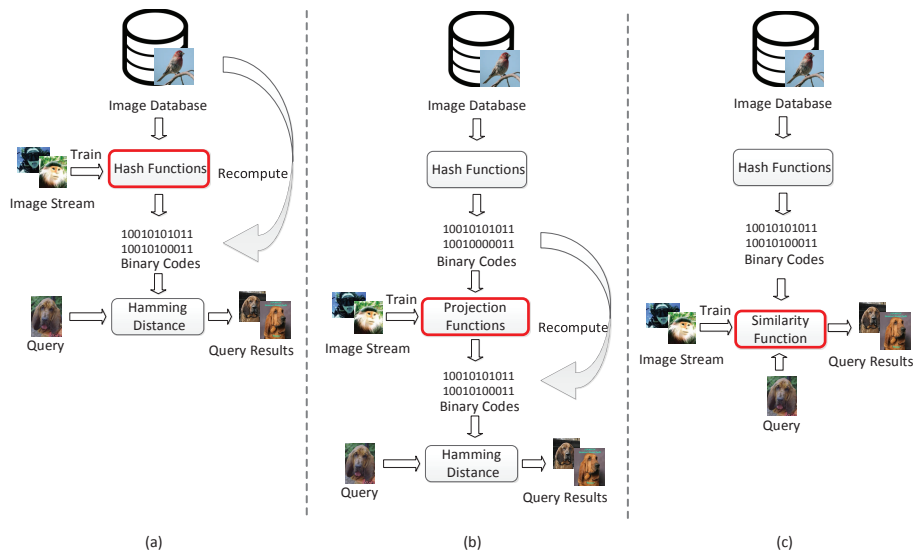}
\caption{Comparison of online hashing frameworks. (a) The framework updates the hash functions and recomputes the binary codes. (b) The framework updates the projections functions and projects the binary codes. (c) The framework updates the similarity function without updating binary codes.}
\label{fig:res0}
\end{figure*}

Locality Sensitive Hashing (LSH)~\cite{andoni2006near} is a representative of the hashing methods. By generating as hash functions random projection vectors sampled from a multidimensional Gaussian distribution with zero mean and identity covariance matrix, LSH can hash similar data into similar binary codes in a high probability. However, LSH is a data-independent hashing method and usually needs long binary codes to achieve a good precision. To obtain compact binary codes for the data, data-dependent hashing methods~\cite{weiss2009spectral,weiss2012multidimensional,he2019k,weng2020concatenation} employ machine learning techniques to learn the hash functions according to the data. For example, some hashing methods~\cite{ge2014graph,liu2014discrete,AAAI1714686} build graphs to describe the relationship between the data and learn the hash functions from the graphs. Some hashing methods~\cite{gong2013iterative,he2013k} adopt dimensionality reduction techniques to project the data into a low-dimensional space and quantize each dimension into a bit. Due to the development of deep learning methods, deep hashing methods~\cite{shen2018unsupervised,do2016learning,cao2018deep,wu2019deep} learn the neural networks as the hash functions. Since the memory space is limited, deep hashing methods learn the hash functions by using minibatch-based stochastic optimization which requires multiple passes over a given database. Although these methods can achieve good search performance, they are designed for the static database and have poor scalability for the ever-growing database. When new data arrives, the hash functions have to be re-learnt by accumulating all the training data, which is time-consuming.

To address this challenge, online hashing methods~\cite{DBLP:conf/aaai/LinJLSWW19,Cakir2017MIHash,lin2020fast,lin2018supervised,weng2019fast} employ online learning techniques to learn the hash functions online from the streaming data. For example, Online Sketching Hashing (OSH)~\cite{leng2015online} maintains a small-size data sketch from the streaming data and learns hash functions from the data sketch. Online Kernel Hashing (OKH)~\cite{7907165} constructs a loss function by using the label information and updates the hash functions according to the loss function. Balanced Similarity for Online Discrete Hashing (BSODH)~\cite{DBLP:conf/aaai/LinJLSWW19} uses an asymmetric graph regularization to preserve the similarity between the streaming data and the existing dataset and update the hash function according to the graph. Online Supervised Hashing (OSupH)~\cite{cakir2017online} and Hadamard Matrix Guided Online Hashing (HMOH)~\cite{Lin2020} explore how to generate the target codes according to the label information and update the hash functions according to the target codes. Although the current online hashing methods can achieve good search performance and are efficient in learning the hash functions online, they have to recompute the binary codes for the database when the hash functions are updated. It is inefficient since recomputing the binary codes needs to accumulate the whole database as shown in Fig.~\ref{fig:res0}(a). Mutual Information Hashing (MIH)~\cite{Cakir2017MIHash} has noticed this problem and introduces a trigger update module to reduce the frequency of recomputing the binary codes, but the calculation in the trigger update module is time-consuming~\cite{AAAI2020online}. Online Hashing with Efficient Updating (OHWEU)~\cite{AAAI2020online} tries to solve this problem from another perspective. It designs an online hashing framework that fixes the hash functions and learns the projection functions online to project the binary codes into another binary space. Hence, OHWEU can update the binary codes efficiently without accumulating the whole database, which is shown in Fig.~\ref{fig:res0}(b). However, for OHWEU, the cost of updating the binary codes is increasing as data is growing.

In this paper, we propose a new online hashing framework without updating binary codes. In this framework, we fix the hash functions and introduce a similarity function to measure the similarity between the query and the binary codes. And the similarity function is learnt online from the streaming data to adapt to the data variations. Fig.~\ref{fig:res0} shows the difference between our framework (Fig.~\ref{fig:res0}(c)) and other online hashing frameworks. Compared with the other two online hashing framework, our framework omits the process of updating the binary codes, which is efficient for the ever-growing data. Our contribution in this paper is three fold:

\begin{itemize}
  \item A novel online hashing framework without updating binary codes is proposed by introducing a parametric similarity function that is learnt online from the streaming data.
  \item A new online metric learning algorithm is proposed to learn the similarity function that has a bilinear form, which can treat the query and the binary codes asymmetrically.
  \item The experiments on two multi-label datesets show that our method not only achieves competitive or higher search accuracy than other online hashing methods but also is more efficient in the online process.
\end{itemize}
\section{Related Work}

\subsection{Hashing methods with similarity functions}
Since Hamming distance has a limited distance range and easily causes the ambiguity problem, some emerging studies~\cite{ji2014query,zhang2013binary,Duan2015Weighted,8288679} explore using other similarity functions to measure the similarity between binary codes rather than Hamming distance. For example, Manhattan  Hashing~\cite{kong2012manhattan} uses Manhattan distance to replace Hamming distance as Manhattan distance can preserve more information about the neighborhood structure in the original feature than Hamming distance. And some studies~\cite{Duan2015Weighted,8288679,weng2016asymmetric} adopt weighted Hamming distance to replace Hamming distance and allocate the weights to each bit.

Although Manhattan distance and weighted Hamming distance can provide superior accuracy than Hamming distance, the computation of Manhattan distance and weighted Hamming distance is slower than that of Hamming distance as the search with Hamming distance can be accelerated in a non-exhaustive search way which can provide a sub-linear search time. To solve this problem, recently, Multi-Index Weighted Querying (MIWQ)~\cite{AAAI2020query} develops a non-exhaustive search method to accelerate the search with weighted Hamming distance, propelling the development of the hashing methods with weighted Hamming distance.

In this paper, our method uses a bilinear similarity function to replace Hamming distance since the similarity function can be learnt in an online way to adapt to the data variations in the streaming environment. Meanwhile, the computation of the bilinear similarity between the binary codes can be represented in a form of weighted Hamming distance, so that it can be accelerated by MIWQ.

\subsection{Metric learning methods}
Metric learning methods~\cite{kulis2012metric,wang2015survey} aim to learn a similarity function tuned to a particular task and are widely used in different applications that rely on distances or similarities. In terms of learning methodology, metric learning methods can be categorized as batch-based learning methods~\cite{wang2017deep,cakir2019deep} and online learning methods~\cite{gao2017sparse,Chechik:2010:LSO:1756006.1756042}.

Our method follows the online-learning methodology. Our method is related to LogDet Exact Gradient Online (LEGO)~\cite{jain2009online} which applies online metric learning technique on updating the hash functions generated by LSH. However, in LEGO, the binary codes need to be recomputed when the hash functions are updated, which is the main difference between LEGO and our method.

\section{Online Hashing with Similarity Learning}
\subsection{Similarity function}
Hashing methods learn the hash functions to map a $D$-dimensional feature vector ${\bf{x}} \in {{\mathbb R}^D}$ to a $b$-bit binary hash code ${\bf{\tilde x}} \in {\{  - 1,1\} ^b}$. Given a query ${\bf{q}} \in {{\mathbb R}^D}$, the query $\bold{q}$ is hashed into a binary hash code $\bold{\tilde q} \in {\{  - 1,1\} ^b}$ and Hamming distance between the binary query $\bold{\tilde q}$ and the binary code ${\bf{\tilde x}}$ is calculated as
\begin{equation}
{S_H}({\bf{\tilde q}},{\bf{\tilde x}}) = \sum\limits_{i = 1}^b {{{\tilde q}_i} \otimes {{\tilde x}_i}},
\label{alg1}
\end{equation}
where $\otimes$ is an xor operation.

We consider a parametric similarity function that has a bilinear form to replace Hamming distance, which is,
\begin{equation}
{S_{{\bf{\tilde M}}}}({\bf{\tilde q}},{\bf{\tilde x}}) = {{\bf{\tilde q}}^T}{\bf{\tilde M\tilde x}},
\label{alg2}
\end{equation}
where ${\bf{\tilde M}} \in {\mathbb R}^{b \times b}$.

There are two advantages to use the bilinear similarity: (a) It can avoid imposing positivity or square constraints that result in computationally intensive solutions. (b) It can be combined with MIWQ~\cite{AAAI2020query} to accelerate the search process. These two advantages are elaborated in the following.

In~\cite{NIPS2013_5017,AAAI2020online},  it shows that treating the queries and the database points asymmetrically can improve the search accuracy. In detail, although the database points are hashed into binary codes, the query does not need to be hashed into a binary code so that more information of the query can be preserved. Hence, we treat the query points and the database points in an asymmetric way by modifying Eqn. (\ref{alg2}) as
\begin{equation}
{S_{\bf{M}}}({\bf{q}},{\bf{\tilde x}}) = {{\bf{q}}^T}{\bf{M\tilde x}},
\label{alg3}
\end{equation}
where ${\bf{M}} \in {\mathbb R}^{D \times b}$, $\bf q$ is the original feature vector and is not mapped to a binary code.

Since the query is fixed in each comparison between the query and the binary codes in the process of searching for the nearest neighbors of the query, the multiplication between q and M can be pre-computed and stored. Hence, Eqn. (3) is rewritten as
\begin{equation}
{S_{\bf{M}}}({\bf{q}},{\bf{\tilde x}}) = {{\bf{\hat m}}^T}{\bf{\tilde x}},
\end{equation}
where ${\bf{\hat m}} = {({{\bf{q}}^T}{\bf{M}})^T}$.

Further, Eqn. (4) can be rewritten as a form of weighted Hamming distance
\begin{equation}
{S_{\bf{M}}}({\bf{q}},{\bf{\tilde x}}) = \sum\limits_{i = 1}^b {{w_i}({{\tilde x}_i})} ,
\end{equation}
where ${\tilde x}_i$ is the $i^{th}$ bit of $\bf{\tilde x}$, ${w_i}:\{  - 1,1\}  \to {\mathbb R}$ is a function to store the pre-computed value for the $i^{th}$ bit and is defined as
\begin{equation}
\left\{ \begin{array}{l}
{w_i}( - 1) =  - {{\hat m}_i}\\
{w_i}(1) = {{\hat m}_i}.
\end{array} \right.
\end{equation}
and ${{\hat m}_i}$ is the $i^{th}$ element of ${\bf{\hat m}}$.

Since the computation of the similarity function in Eqn. (3) can be rewritten as a form of weighted Hamming distance Eqn. (5), it can be combined with MIWQ~\cite{AAAI2020query} to perform the non-exhaustive search with the learnt similarity when given a query. In~\cite{AAAI2020query}, it shows that the non-exhaustive search with weighted Hamming distance is competitive to the search with Hamming distance for the compact binary codes in terms of search efficiency. More details of MIWQ can be seen in~\cite{AAAI2020query}.

In our framework, the hash functions are learnt in advance and fixed. Following other online hashing methods, the linear projection-based hash functions are adopted, which are
\begin{equation}
{\bf{\tilde x}} = {\mathop{\rm sgn}} ({{\bf{W}}^T}{\bf{x}} + {\bf{b}}),
\label{alg7}
\end{equation}
where $\bold{W} \in {\mathbb R}^{D \times b}$ is a projection matrix and $\bold{b} \in {\mathbb R}^{b}$ is a threshold vector.

As~\cite{7907165,AAAI2020online} do, by assuming there is a small amount of training data in the initial stage, we can use the training data to learn the hash functions that can preserve the data similarity in the initial stage. Here, we learn the hash functions by using Iterative Quantization (PCA-ITQ)~\cite{gong2013iterative}  which is an unsupervised hashing method to preserve the similarity between data effectively
for both the seen and the unseen data. It is justified in~\cite{7936671} that the hash functions can be learnt by PCA-ITQ just with a small amount of training data.

\subsection{Online learning}
Although Online Algorithm for Scalable Image Similarity learning (OASIS)~\cite{chechik2009online,Chechik:2010:LSO:1756006.1756042} provides a solution for learning the similarity matrix $\bold{M}$ in Eqn. (3), the similarity matrix $\bold{M}$ in OASIS is a square matrix while it can be a rectangular matrix in our method. Hence, we propose a new online metric learning algorithm to learn the similarity matrix.

The similarity matrix $\bf{M}$ in Eqn. (\ref{alg3}) can be decomposed as
\begin{equation}
\bold{M}=\bold{U}^T\bold{V}
\label{alg8}
\end{equation}
where $\bold{U} \in {\mathbb R}^{l \times D}$ and $\bold{V} \in {\mathbb R}^{l \times b}$ are linear transformation.

Therefore, Eqn. (\ref{alg3}) can be rewritten as
\begin{equation}
{S_{\bf{M}}}({\bf{q}},{\bf{\tilde x}}) = {{\bf{q}}^T}{\bf{M\tilde x}} = {{\bf{q}}^T}{{\bf{U}}^T}{\bf{V\tilde x}} = {({\bf{Uq}})^T}{\bf{V\tilde x}}.
\label{alg9}
\end{equation}

\begin{figure}[!htbp]
\centering

 \includegraphics[width=6cm]{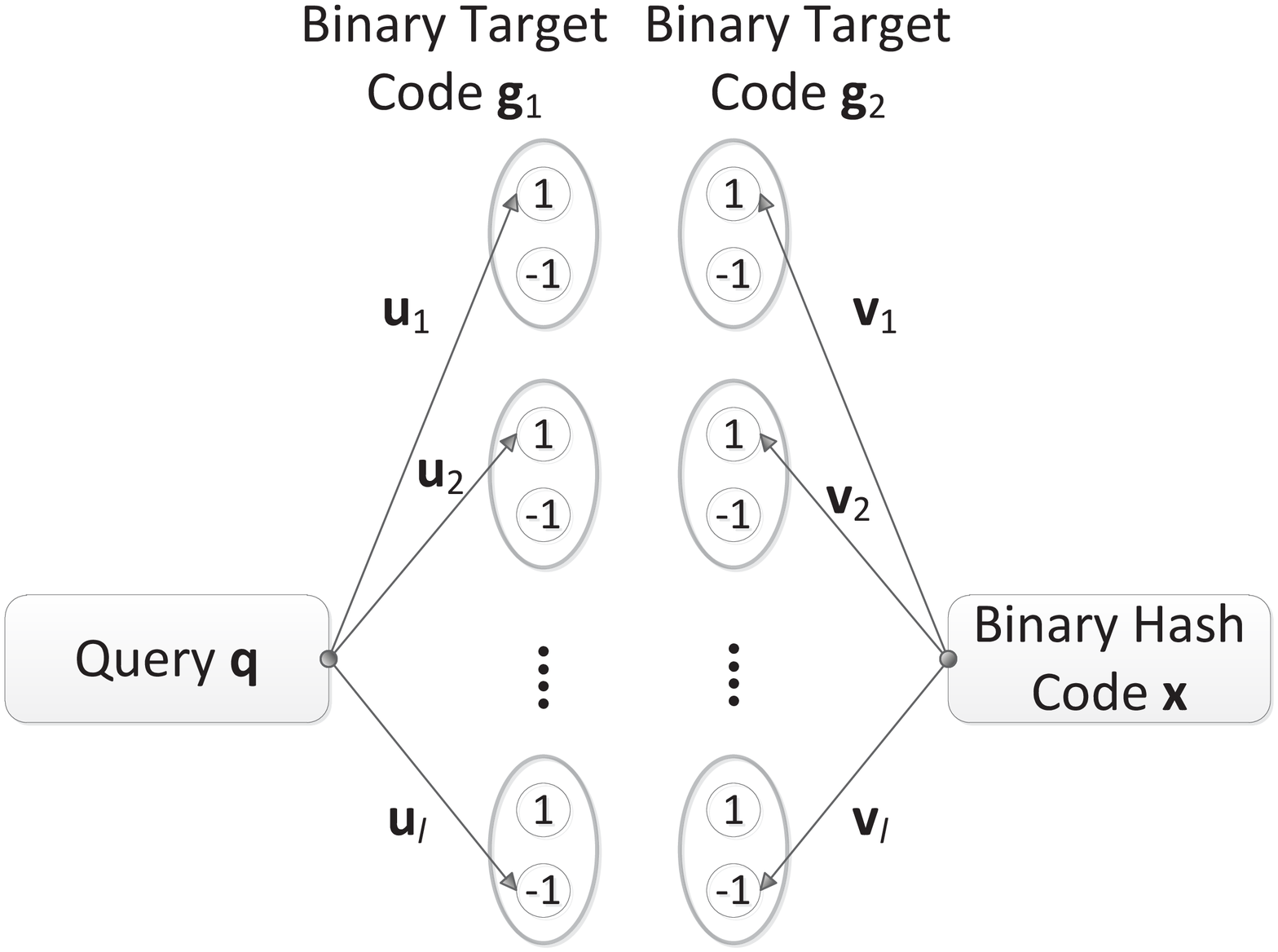}
\caption{An example of learning projection vectors.}
\label{fig:res2}
\end{figure}

The similarity between $\bf{q}$ and $\bold{\tilde x}$ can be treated as the inner product of two vectors ${\bf{Uq}}$ and ${\bf{V\tilde x}}$ . If two data points are similar, the inner product of their corresponding vectors after linear transformation should be larger than that of dissimilar ones. To learn the similarity function using the point-wise label information, inspired by HMOH~\cite{Lin2020}, we can generate $l$-bit binary target code ${\bf{g}} \in {\{  - 1,1\} ^l}$ according to the label information so that each column $\bold{u} \in {\mathbb R}^D$ in $\bold{U}$ is a projection vector and corresponds to one bit of the binary target code. So is each column $\bold{v} \in {\mathbb R}^b$ in $\bold{V}$. This idea is shown in Fig.~\ref{fig:res2}.

It should be noted that the length $l$ of the binary target code ${\bf{g}}$ will affect the performance of our method. When $l$ increases, the search accuracy of our method usually improves in the case of low bits. Considering about the constraint of the rank of $\bf M$ and the learning efficiency, we take $l=3b$ ($i.e.$, the binary target code is three times the length of the generated binary hash codes) which will be demonstrated in the experiments.

Generating the $l$-bit binary target codes that can preserve the label information has been explored in some online hashing methods~\cite{cakir2017online,9126268,Lin2020,Lin:2018:SOH:3240508.3240519}. Here, we adopt HMOH~\cite{Lin2020} to introduce the Hadamard matrix and regard each column of Hadamard matrix as the binary target code ${\bf{g}} \in {\{  - 1,1\} ^l}$ for each class label, which by nature satisfies several desired properties of binary hash codes.

As shown in Fig.~\ref{fig:res2}, after obtaining the binary target codes according to the label information, each projection vector $\bf u$ in $\bf U$ can be learnt online as an online binary classification problem. We adopt the Passive-Aggressive algorithm~\cite{crammer2006online} to learn the projection vector as it can converge to a high quality projection vector after learning from a small amount of training data.

When a data point $\bf x$ with its binary target code $\bf g$ arrives, for each projection vector $\bf u$ and its corresponding bit $g$ in the binary target code $\bf g$, the hinge-loss function is defined as
\begin{equation}
\ell ({\bf{u}};({\bf{x}},g)) = \left\{ \begin{array}{l}
0\quad \quad \quad \quad \quad\quad\;g({{\bf{u}}^T}{\bf{x}}) \ge 1\\
1 - g({{\bf{u}}^T}{\bf{x}})\;\;\,\quad otherwise.
\end{array} \right.
\label{alg11}
\end{equation}

For brevity, we use $\ell_i$ to represent $\ell ({\bf{u}};({{\bf{x}}_i},{g_i}))$  where $\bold{x}_i$ is the data point and $g_i$ is the corresponding bit of the binary target code $\bold{g}_i$ in the $i^{th}$ round. At first, $\bf u$ is initialized to a zero-valued vector. Then, for each round $i$, we solve the following convex problem with soft margin:
\begin{equation}
\begin{array}{l}
{{\bf{u}}_i} = \mathop {\arg \;\min }\limits_{\bf{u}} \;{\textstyle{1 \over 2}}||{\bf{u}} - {{\bf{u}}_{i - 1}}|{|^2} + C\xi \\
s.t.\quad {\ell _i} \le \xi \;and\;\xi  \ge 0
\end{array}
\label{alg12}
\end{equation}
where $||\cdot||$ is the Euclidean norm and $C$ controls the trade-off between making $\bf u$ stay close to the previous parameter $\bold{u}_{i-1}$ and minimizing the loss on the current loss $\ell_i$.

When $\ell_i$ = 0, $\bold{u}_i=\bold{u}_{i-1}$ satisfies Eqn.(\ref{alg12}) directly. Otherwise, we define the Lagrangian as:
\begin{equation}
L(\bf{p},\tau,\lambda,\xi) = {\textstyle{1 \over 2}}||{\bf{p}} - {{\bf{p}}_{i - 1}}||^2 + C\xi + \tau (1-\xi-{g_i^*}({{\bf{p}}^T}{{\bf{h}}_i})) - \lambda\xi,
\label{alg13}
\end{equation}
with $\tau \ge 0$ and  $\lambda \ge 0$ are Lagrange multipliers.

Let $\frac{{\partial L({\bf{u}},\tau ,\lambda ,\xi )}}{{\partial {\bf{u}}}} = {\bf{u}} - {{\bf{u}}_{i - 1}} - \tau {g_i}{{\bf{x}}_i} = 0$, we obtain
\begin{equation}
{\bf{u}} = {{\bf{u}}_{i - 1}} + \tau {g_i}{{\bf{x}}_i}.
\label{alg14}
\end{equation}

Let $\frac{{\partial L({\bf{u}},\tau ,\lambda ,\xi )}}{{\partial \xi }} = C - \tau  - \lambda  = 0$, we obtain
\begin{equation}
C = \tau  + \lambda,
\label{alg15}
\end{equation}
where $\tau  \le C$ as $\lambda  \ge 0$

Plugging Eqn. (14) and (15) back into Eqn. (\ref{alg13}), we obtain $L(\tau ) =  - {\textstyle{1 \over 2}}{\tau ^2}||{{\bf{x}}_i}|{|^2} + \tau (1 - {g_i}({\bf{u}}_{i - 1}^T{{\bf{x}}_i}))$. Let $\frac{{\partial L(\tau )}}{{\partial \tau }} =  - \tau ||{{\bf{x}}_i}|{|^2} + (1 - {g_i}({\bf{u}}_{i - 1}^T{{\bf{x}}_i})) = 0$, we obtain
\begin{equation}
\tau  = \frac{{1 - {g_i}({\bf{u}}_{i - 1}^T{{\bf{x}}_i})}}{{||{{\bf{x}}_i}||}}.
\label{alg16}
\end{equation}

Therefore, the solution to Eqn. (\ref{alg12}) is
\begin{equation}
{{\bf{u}}_i} = {{\bf{u}}_{i - 1}} + \tau {g_i}{{\bf{x}}_i},
\label{alg17}
\end{equation}
where
\begin{equation}
\tau  = \min \{ C,\frac{{1 - {g_i}({\bf{u}}_{i - 1}^T{{\bf{x}}_i})}}{{||{{\bf{x}}_i}||}}\}.
\label{alg18}
\end{equation}

Similarly, for each projection vector $\bold{v}$, the data point $\bf x$ is hashed into a binary hash code $\bf \tilde x$ by hash functions in Eqn. (\ref{alg7}), and the hinge-loss function is defined as
\begin{equation}
\tilde \ell ({\bf{v}};({\bf{\tilde x}},g)) = \left\{ \begin{array}{l}
0\quad \quad \quad \quad \quad\quad\;g({{\bf{v}}^T}{\bf{\tilde x}}) \ge 1\\
1 - g({{\bf{v}}^T}{\bf{\tilde x}})\quad\;\;\,otherwise.
\end{array} \right.
\label{alg19}
\end{equation}

For brevity, we use $\tilde \ell_i$ to represent $\tilde \ell ({\bf{v}};({\bf{\tilde x_i}},g_i))$. At first, $\bf v$ is initialized to a zero-valued vector. Then, for each round $i$, we solve the following convex problem with soft margin:
\begin{equation}
\begin{array}{l}
{{\bf{v}}_i} = \mathop {\arg \;\min }\limits_{\bf{v}} \;{\textstyle{1 \over 2}}||{\bf{v}} - {{\bf{v}}_{i - 1}}|{|^2} + C\xi \\
s.t.\quad {\tilde \ell _i} \le \xi \;and\;\xi  \ge 0.
\end{array}
\label{alg20}
\end{equation}

Following the above derivation, the solution to Eqn. (\ref{alg20}) is
\begin{equation}
{{\bf{v}}_i} = {{\bf{v}}_{i - 1}} + \tau {g_i}{{\bf{\tilde x}}_i},
\label{alg21}
\end{equation}
where
\begin{equation}
\tau  = \min \{ C,\frac{{1 - {g_i}({\bf{v}}_{i - 1}^T{{\bf{\tilde x}}_i})}}{{||{{\bf{\tilde x}}_i}||}}\}.
\label{alg22}
\end{equation}

The pseudo-code of our method, Online Hashing with Similarity Learning (OHSL), is summarized in Algorithm 1.
\begin{algorithm}
\caption{OHSL}               
\label{algorithm1}                      
\begin{algorithmic}[1]
\REQUIRE streaming data $(\bold{x}_i,\bold{y}_i)$, parameters $C$ and $l$               
\ENSURE $\bold{M}$  
\STATE Initialize $\bold{U}=[0,...,0]$ and $\bold{V}=[0,...,0]$
\FOR {$i=1,2,...$}
\STATE Obtain the binary target code $\bold{g}_i$ following HMOH
\FOR {each $\bf u$ in $\bf U$}
\STATE Update $\bf u$ by Eqn. (\ref{alg17})
\ENDFOR
\STATE Obtain the binary representation $\bold{\tilde x}_i$ by Eqn.(\ref{alg7})
\FOR {each $\bf v$ in $\bf V$}
\STATE Update $\bf v$ by Eqn. (\ref{alg21})
\ENDFOR
\STATE $\bold{M} = \bold{U}^T\bold{V}$
\ENDFOR

\end{algorithmic}
\end{algorithm}

\section{Experiments}
\subsection{Datasets}
The following experiments are performed on two commonly-used multi-label image datasets, MS-COCO and NUS-WIDE.

(a).MS-COCO dataset~\cite{10.1007/978-3-319-10602-1_48}. The MS-COCO dataset is a multi-label dataset including training images and validation images. Each image is labeled by some of the 80 concepts. After filtering out the images that do not contain any concept label, there are 122,218 images left. Each images is represented by a 4096-D feature extracted from the fc7 layer of a VGG-16 network~\cite{Simonyan14c} pretrained on ImageNet~\cite{imagenet_cvpr09}. For each category, we randomly take 50 images as the query images and the rest as the database images.

(b). NUS-WIDE dataset~\cite{Chua:2009:NRW:1646396.1646452}. The NUS-WIDE dataset is a multi-label dataset including 269,648 web images associated with tags. Following~\cite{Kang:2016:CSB:3015812.3015994}, the images that belong to the 21 most frequent concepts are selected and there are 195,834 images left. Each images is represented by a 4096-D feature extracted from the fc7 layer of a VGG-16 network~\cite{Simonyan14c} pretrained on ImageNet. For each category, we randomly take 50 images as the query images and the rest as the database images.

\begin{table}[!ht]

	\caption{The mAP results on MS-COCO. The best result is labeled with boldface and the second best has an underline.}
	\begin{center}
		\begin{tabular}{|c|c|c|c|c|}\hline
           \multirow{2}{*}{}    & \multicolumn{4}{c|}{mAP}      \\ \cline{2-5}
			                    &  16 bits & 32 bits & 64 bits & 96 bits \\ \hline
           $\bold{OHSL}$        &  $\bold{0.524}$ & $\bold{0.536}$ & $\bold{0.574}$ & $\bold{0.578}$  \\ \hline
            OHWEU	            &  0.444   &  \underline{0.506}	 &   \underline{0.539}	& \underline{0.567} \\ \hline
            HMOH	            &  0.460   &  0.501	 &   0.534	&  0.550 \\ \hline
             MIH	            &  0.442   &  0.435	 &   0.438	&  0.488 \\ \hline
            OSupH	            &  \underline{0.463}   &  0.495	 &   0.512	&  0.514 \\ \hline
             OKH	            &  0.412   &  0.436  &   0.470  &  0.484 \\ \hline
            BSODH	            &  0.458   &  0.437	 &   0.434	&  0.431 \\ \hline
             OSH	            &  0.457   &  0.483	 &   0.499	&  0.504 \\ \hline

		\end{tabular}
	\end{center}
	\label{table0}

\end{table}

\begin{table}[!ht]

	\caption{The mAP results on NUS-WIDE. The best result is labeled with boldface and the second best has an underline.}
	\begin{center}
		\begin{tabular}{|c|c|c|c|c|}\hline
           \multirow{2}{*}{}    & \multicolumn{4}{c|}{mAP}      \\ \cline{2-5}
			                    &  16 bits & 32 bits & 64 bits & 96 bits \\ \hline
           $\bold{OHSL}$&  \underline{0.667} & \underline{0.673}	&\underline{0.700}  &\underline{0.701} \\ \hline
            OHWEU	            &  0.552   &  0.637  &   0.653	&  0.670 \\ \hline
            HMOH	            &  $\bold{0.668}$   &  $\bold{0.716}$  &   $\bold{0.725}$	&  $\bold{0.728}$ \\ \hline
             MIH	            &  0.571   &  0.635	 &   0.649  &  0.647  \\ \hline
            OSupH	            &  0.611   &  0.612	 &   0.631	&  0.640\\ \hline
             OKH	            &  0.506   &  0.550	 &   0.555	&  0.598\\ \hline
            BSODH	            &  0.614   &  0.611  &   0.613  &  0.620 \\ \hline
             OSH	            &  0.466   &  0.489	 &   0.514  &  0.522\\ \hline

		\end{tabular}
	\end{center}
	\label{table1}

\end{table}
\subsection{Accuracy comparison}
Following~\cite{Cakir2017MIHash,DBLP:conf/aaai/LinJLSWW19}, mean Average Precision (mAP) is used to measure the search accuracy of the online hashing methods. If the images and the query share at least one common label, they are defined as a groundtruth neighbor. The results are averaged by repeating the experiments 10 times.

We compare our method, OHSL, with OHWEU~\cite{AAAI2020online}, HMOH~\cite{Lin2020}, BSODH~\cite{DBLP:conf/aaai/LinJLSWW19}, MIH~\cite{Cakir2017MIHash}, OSupH~\cite{cakir2017online}, OKH~\cite{7907165} and OSH~\cite{leng2015online}. The key parameters in each compared method are set as the ones recommended in the corresponding papers except a small amount of parameters are adjusted to be suitable for the datasets. In MIH, the reservoir size is set to 200 to make a good tradeoff between the accuracy and the training cost. Following the operation in OKH and OHWEU, we take 300 data points to train the hash functions in the initial stage.

\begin{figure*}[!htbp]
	\centering
	\begin{tabular}{cccc}
		\includegraphics[width=0.48\columnwidth]{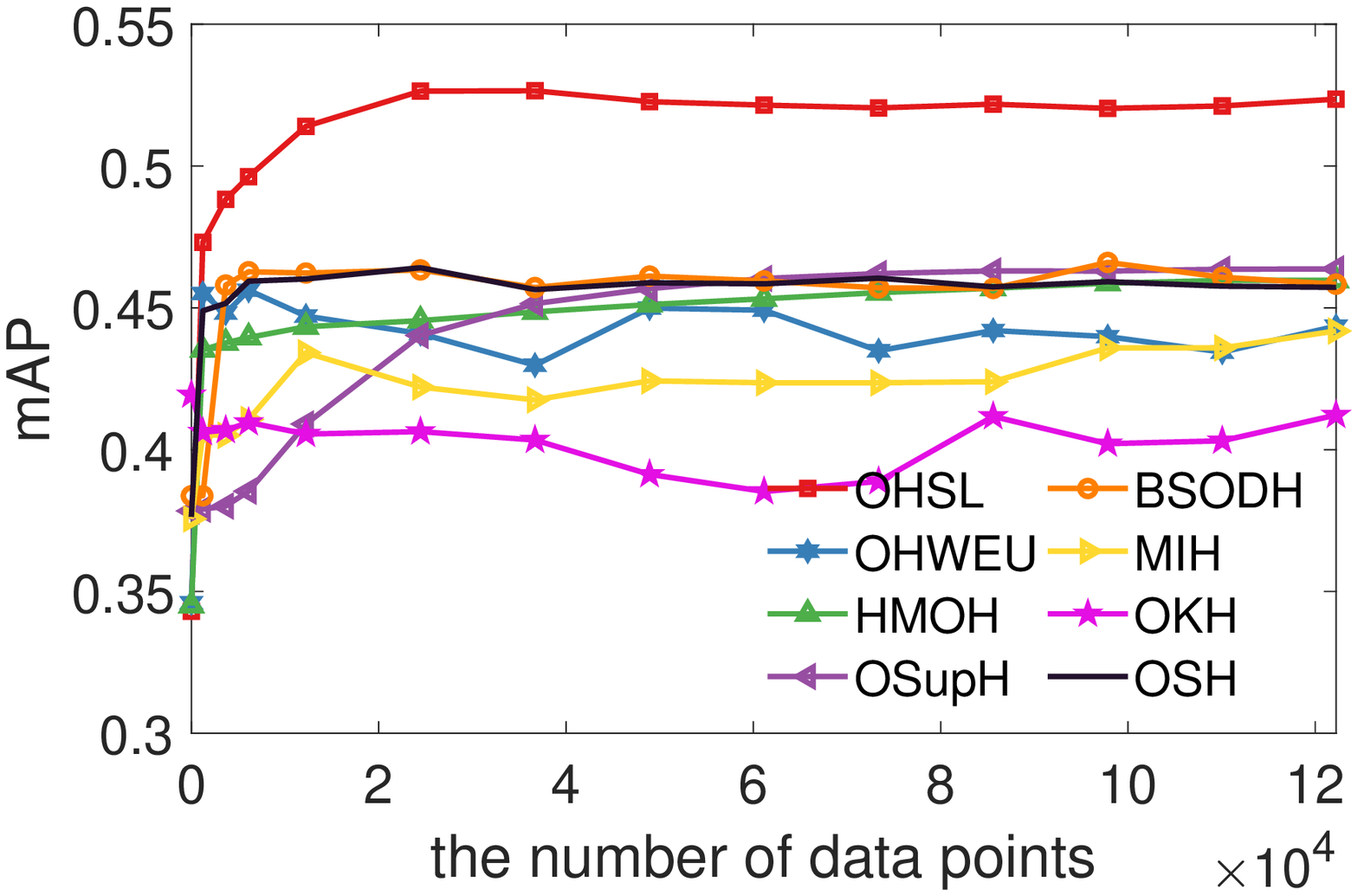}&
		\includegraphics[width=0.48\columnwidth]{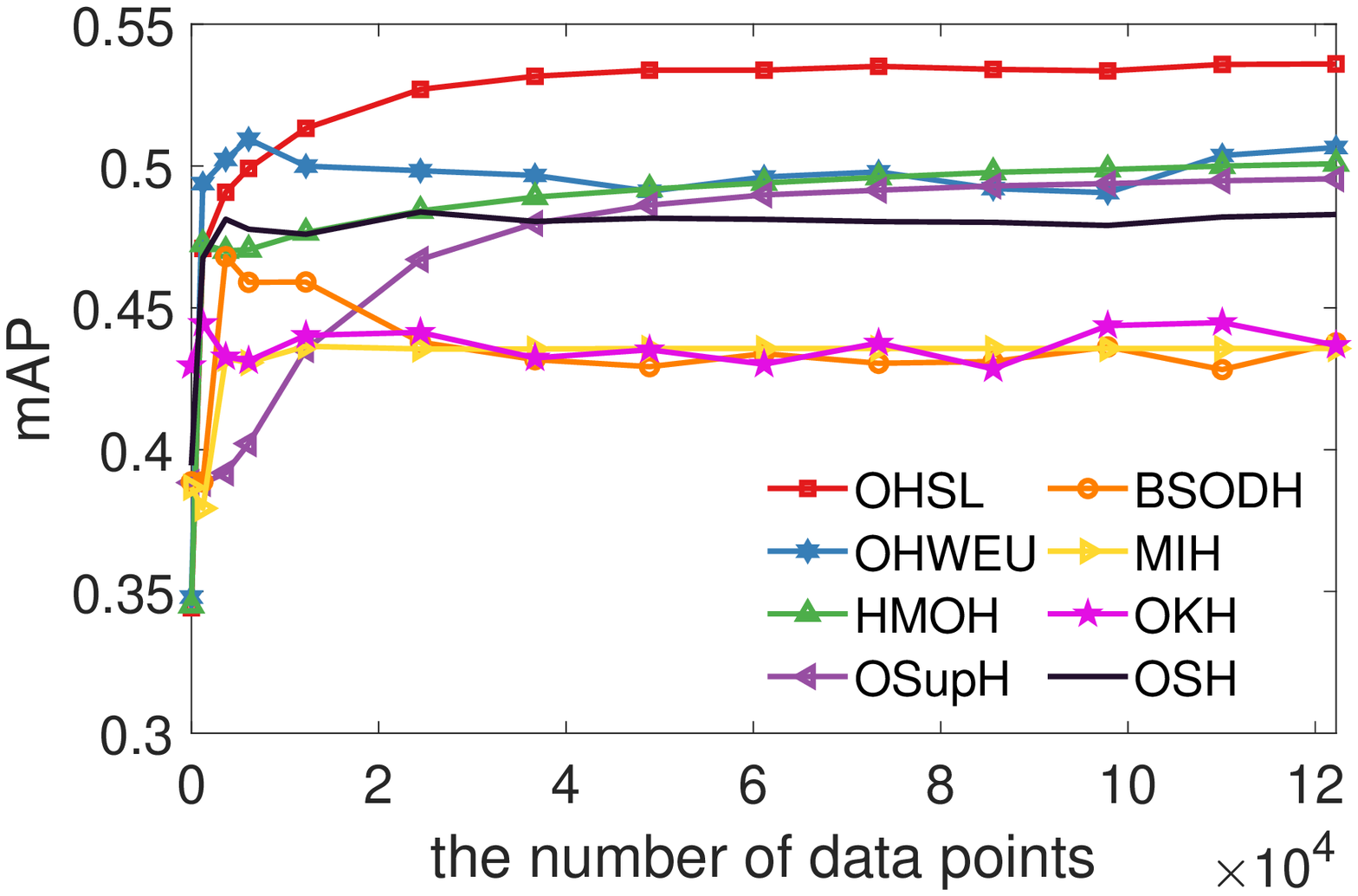}&
		\includegraphics[width=0.48\columnwidth]{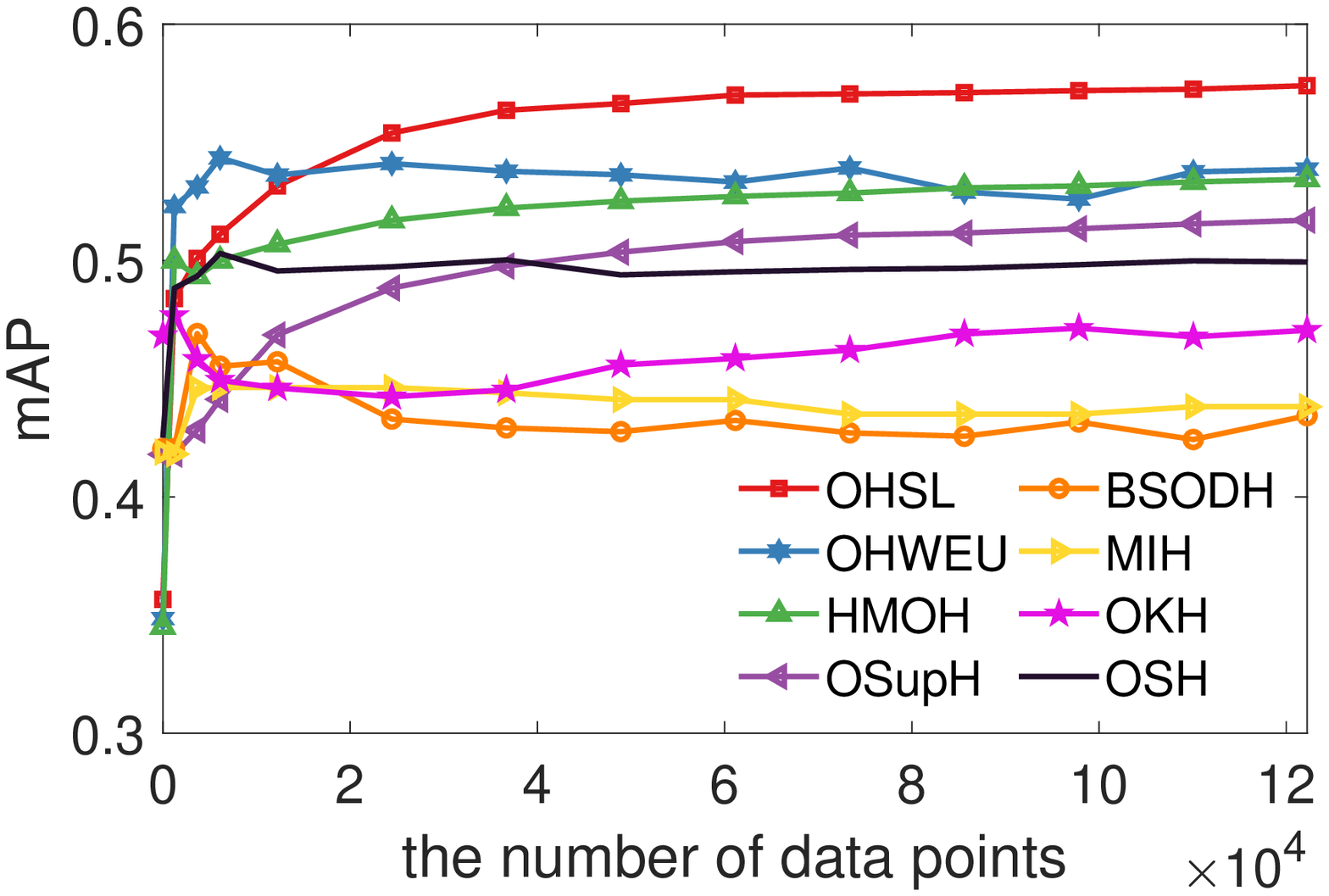}&
		\includegraphics[width=0.48\columnwidth]{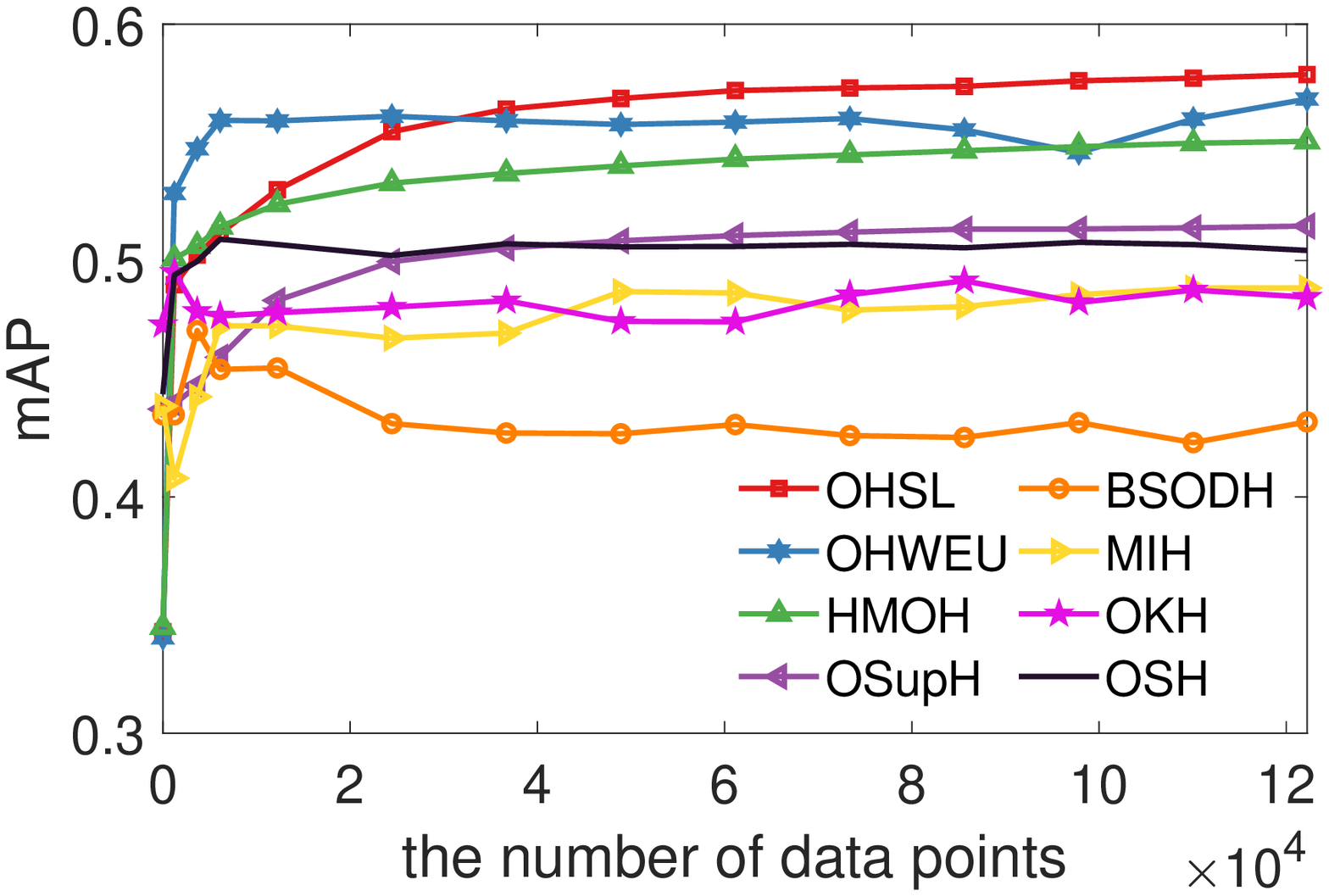}\\
		{\small (a) 16 bits}  &  {\small (b) 32 bits} & {\small (c) 64 bits}  &  {\small (d) 96 bits}
	\end{tabular}
	
	\caption{The mAP results of different online hashing methods with training data increasing on MS-COCO.}
	\label{fig:res3}
\end{figure*}

\begin{figure*}[!htbp]
	\centering
	\begin{tabular}{cccc}
		\includegraphics[width=0.48\columnwidth]{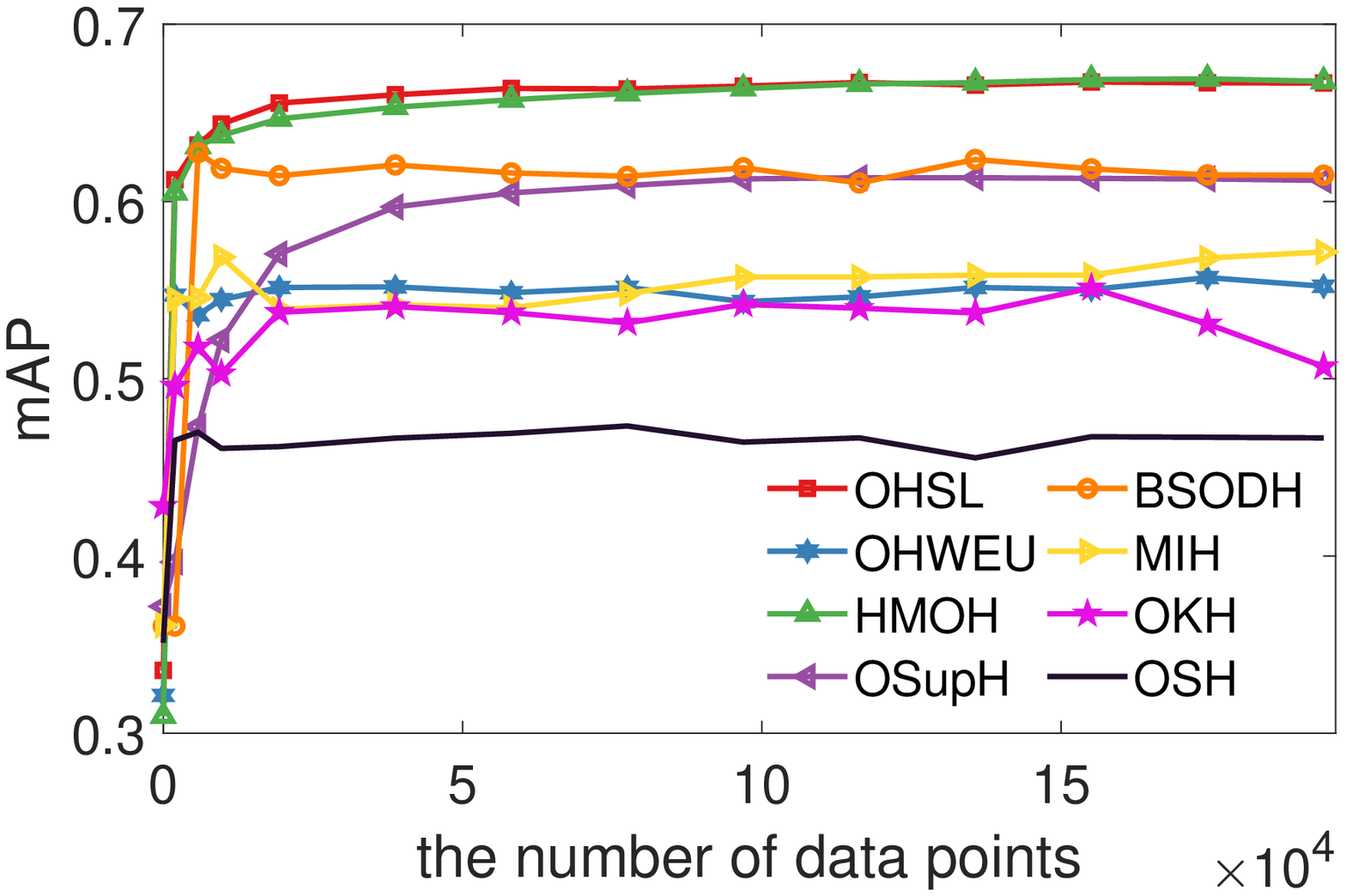}&
		\includegraphics[width=0.48\columnwidth]{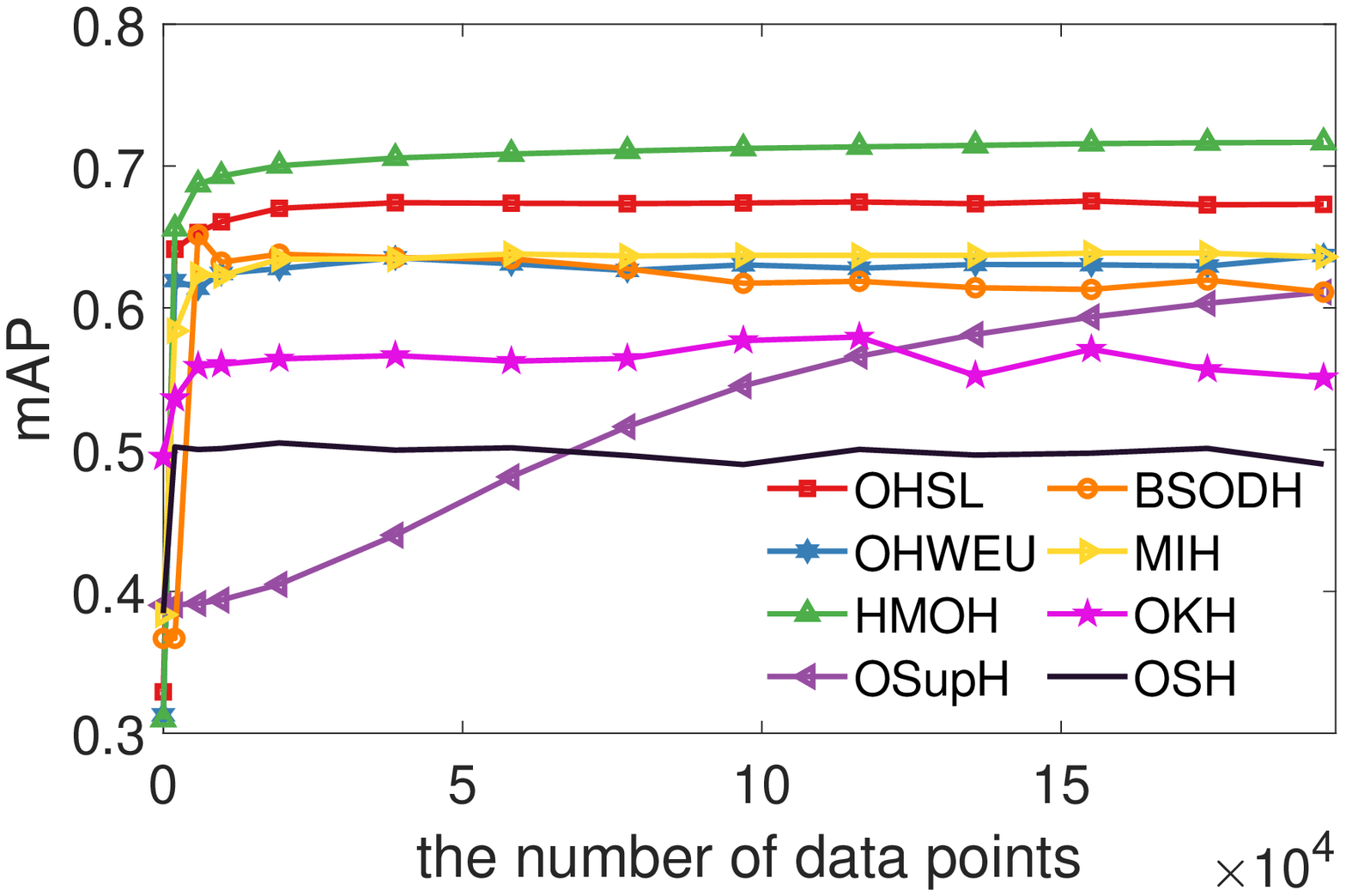}&
		\includegraphics[width=0.48\columnwidth]{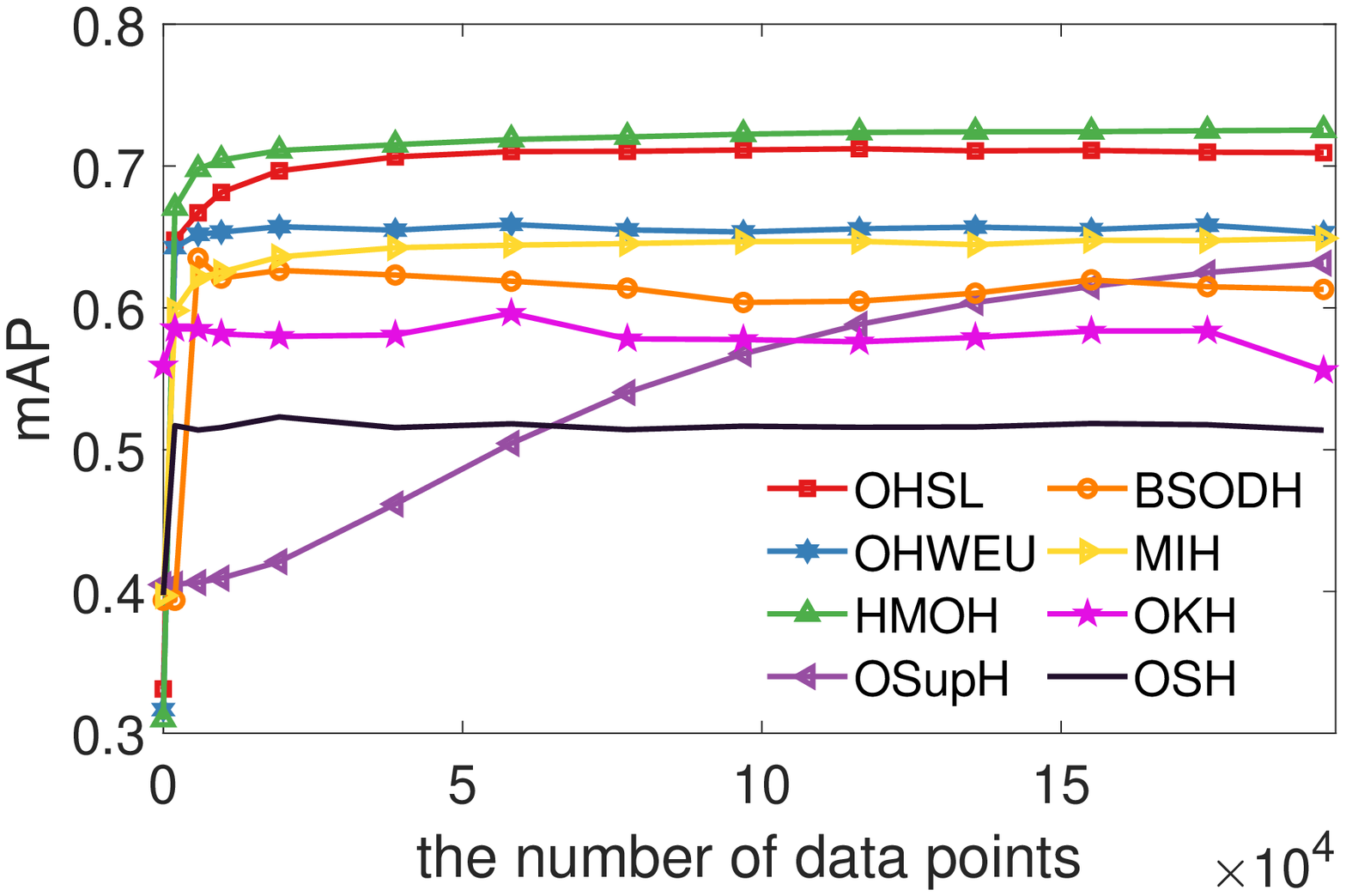}&
		\includegraphics[width=0.48\columnwidth]{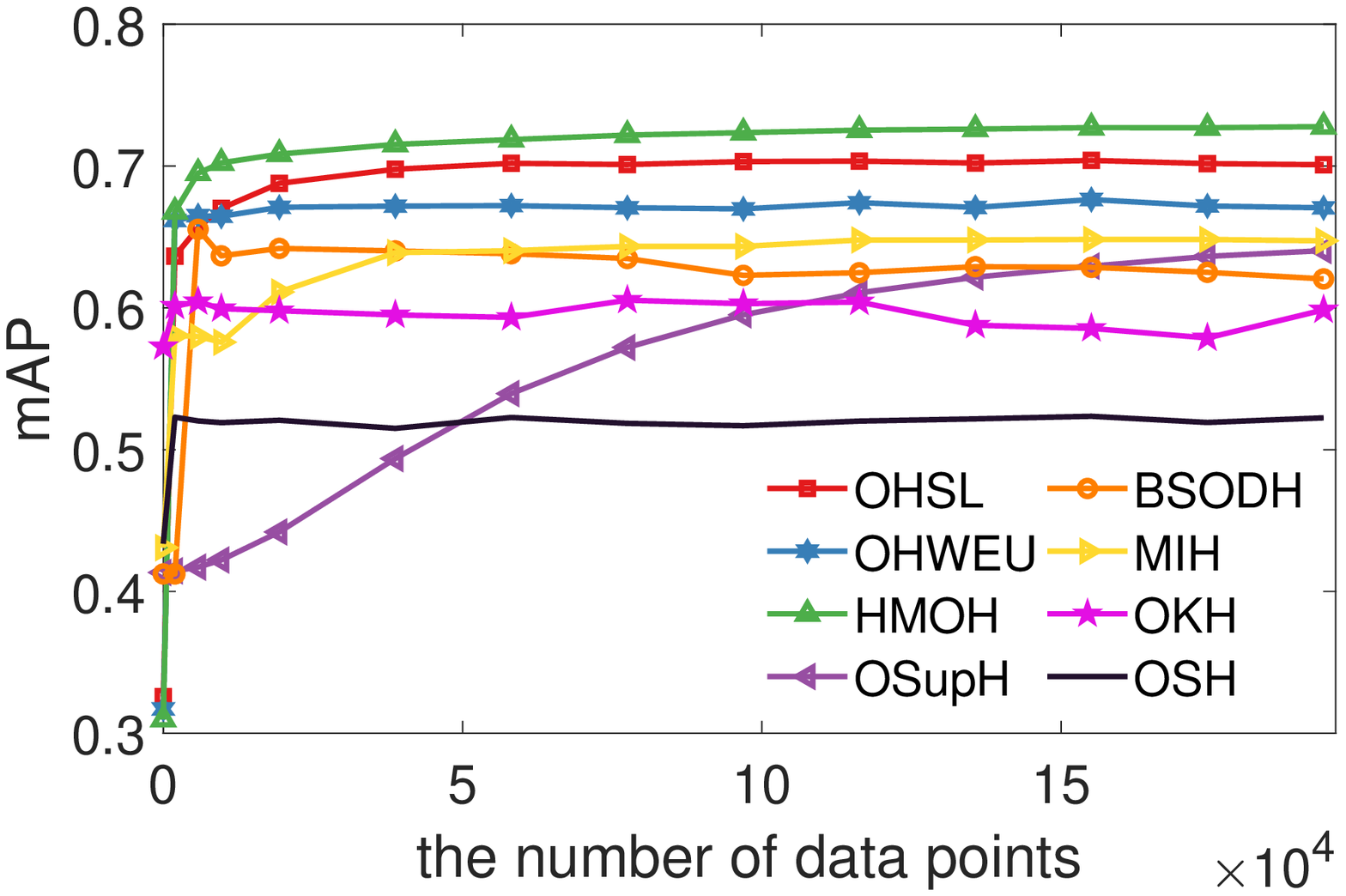}\\
		{\small (a) 16 bits}  &  {\small (b) 32 bits} & {\small (c) 64 bits}  &  {\small (d) 96 bits}
	\end{tabular}
	
	\caption{The mAP results of different online hashing methods with training data increasing on NUS-WIDE.}
	\label{fig:res4}
\end{figure*}

Table~\ref{table0} and Table~\ref{table1} show the mAP results of different online hashing methods on MS-COCO and NUS-WIDE, respectively. The best results are bolded and the second results are underlined. On MS-COCO, OHSL can achieve the best results among the online hashing methods and OHWEU achieves the second best results in most of the cases. On NUS-WIDE, HMOH achieves the best results among the online hashing methods and OHSL achieves the second best results.

Fig.~\ref{fig:res3} shows the mAP results of different online hashing methods with the training data increasing from 16 bits to 96 bits on MS-COCO. From 16 bits to 64 bits, OHSL can achieve the better mAP results than other online hashing methods after taking only a few training data for training. For 96 bits, OHWEU can achieve the better mAP results than other online hashing methods after taking only a few training data for training at first. With the data increasing, OHSL is better than OHWEU.

Fig.~\ref{fig:res4} shows the mAP results of different online hashing methods with the training data increasing from 16 bits to 96 bits on NUS-WIDE. From 16 bits to 96 bits, OHSL and HMOH can achieve the better mAP results than other online hashing methods after taking only a few training data for training. And HMOH is better than OHSL.

\subsection{Efficiency comparison}
To compare the efficiency of different online hashing methods, we calculate the time cost by running the experiments on a PC with Intel i7 3.4 GHz CPU, 24 GB memory. Assume the data comes in chunks~\cite{leng2015online} and each chunk is composed of 1000 data points. The online hashing methods update the functions when receiving one chunk of data.

\begin{figure*}[!htbp]
\centering
\begin{tabular}{c}

 \includegraphics[width=11cm]{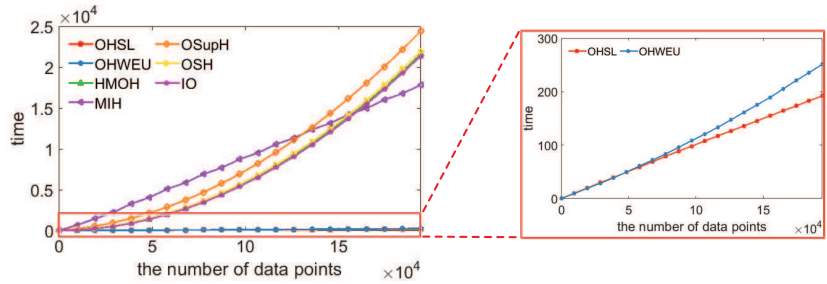}\\

\end{tabular}

\caption{The accumulated time cost of different online hashing methods with data increasing on NUS-WIDE.}
\label{fig:res5}
\end{figure*}

Fig.~\ref{fig:res5} shows the accumulated time cost of different online hashing methods with the data samples increasing on NUS-WIDE for 32 bits. 'IO' denotes the time cost of accumulating all the received data for computing the binary codes by exchanging the data between the hard disk and RAM. The exchanging operation is implemented by C++ code. It takes 3.97s to transfer 1000 4096-D data points from the hard disk to RAM averagely.
For IO, the accumulated time cost rises as a quadratic function of the number of data. For HMOH, OSupH and OSH, the accumulated time cost also rises as a quadratic function of the number of data since they need to take into account the time cost of IO which takes a majority in the total time cost. For MIH, the time cost is approximately linear to the data size since it adopts the trigger update module to reduce the frequency of recomputing the binary codes. However, its time cost is still high due to the time-consuming computations in the trigger update module. Compared with IO, OHSL and OHWEU are much faster, which implicitly denotes that the time cost of exchanging the data is much larger than the time cost of updating the binary codes and learning the functions online. By zooming in the comparison among OHSL and OHWEU, we can see that the accumulated time cost of OHWEU also rises as a quadratic function of the number of data while the accumulated time cost of OHSL is linear to the data size. OHWEU need to update the binary codes once the projection functions are updated and the time cost of each updating operation is linear to the data size. The accumulated time cost of OHWEU rises as a quadratic function of the data size as the accumulated time cost adds up the time cost of each updating operation.

\subsection{Parameter analysis}

There are two key parameters $C$ and $l$ in our method and we investigate the influence of these parameters.

\begin{figure}[!htbp]
\centering
\begin{tabular}{cc}
 \includegraphics[width=0.46\columnwidth]{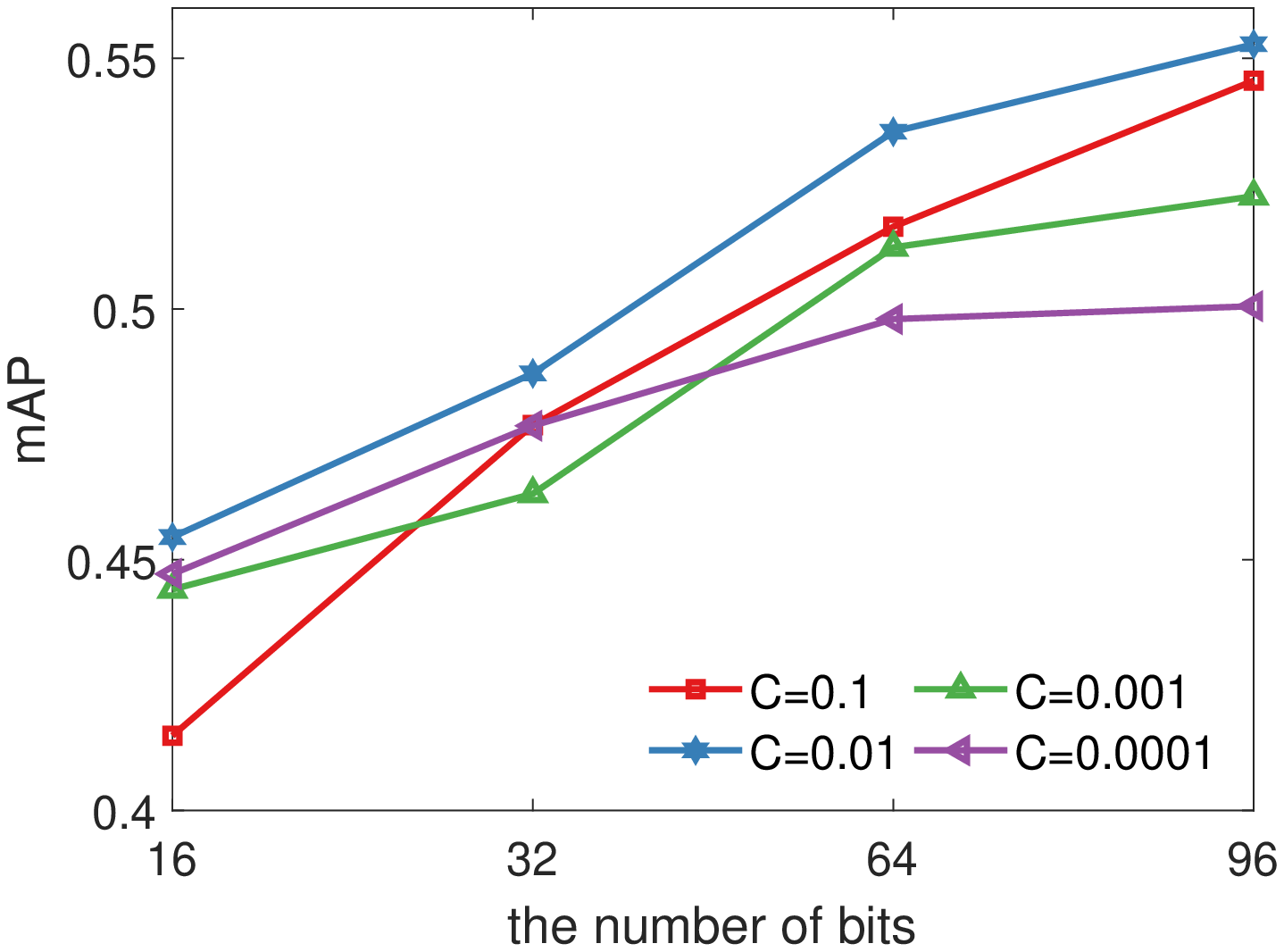}&
 \includegraphics[width=0.46\columnwidth]{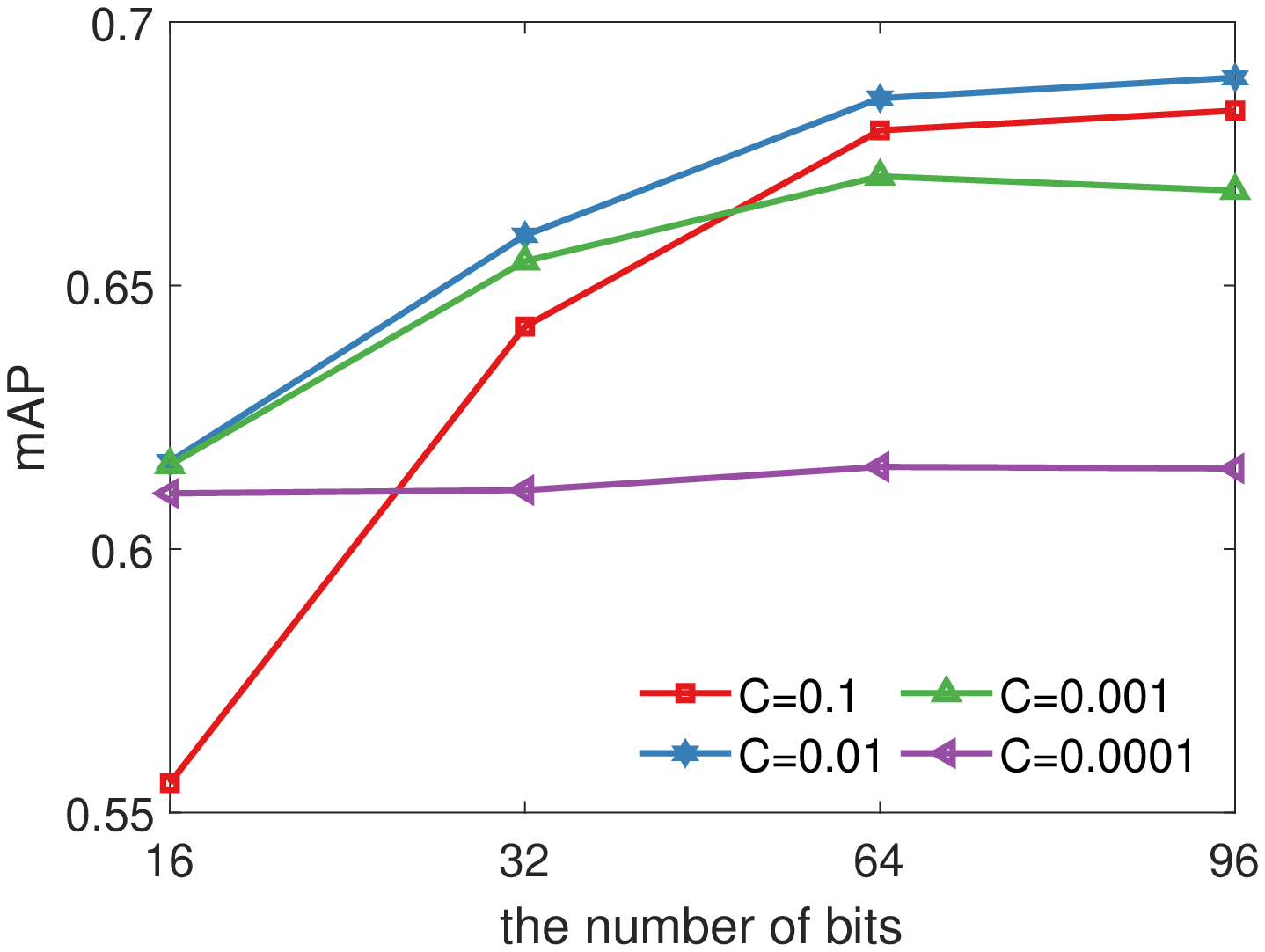}\\
{\small (a) MS-COCO}  &  {\small (b) NUS-WIDE}
\end{tabular}

\caption{The mAP results of different $C$.}
\label{fig:res6}
\end{figure}

By setting the length of the binary target code is equal to the bit number, $i.e.,$ $l=b$, we investigate the influence of the parameter $C$. Fig.~\ref{fig:res6} shows the comparison of OHSL by using different parameter values of $C$. As denoted in Eqn. (11) and (19), $C$ controls the trade-off between making the projection vector stay close to the previous vector and minimizing the current loss. Both $C$ in Eqn. (11) and (19) choose the same value. According to the results, the best parameter choice on MS-COCO and NUS-WIDE is $C = 0.01$.

\begin{figure}[!htbp]
\centering
\begin{tabular}{cc}
 \includegraphics[width=0.46\columnwidth]{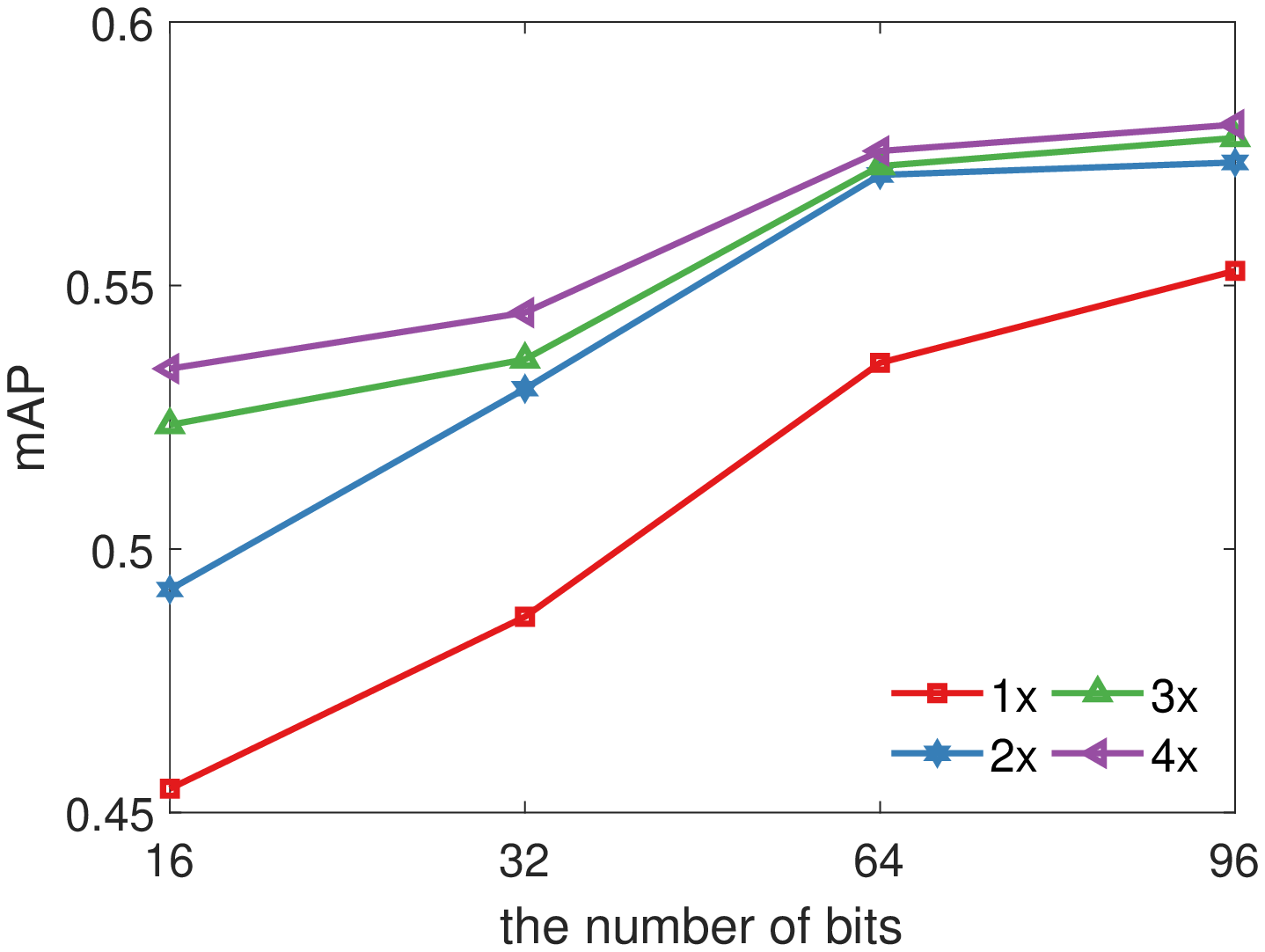}&
 \includegraphics[width=0.46\columnwidth]{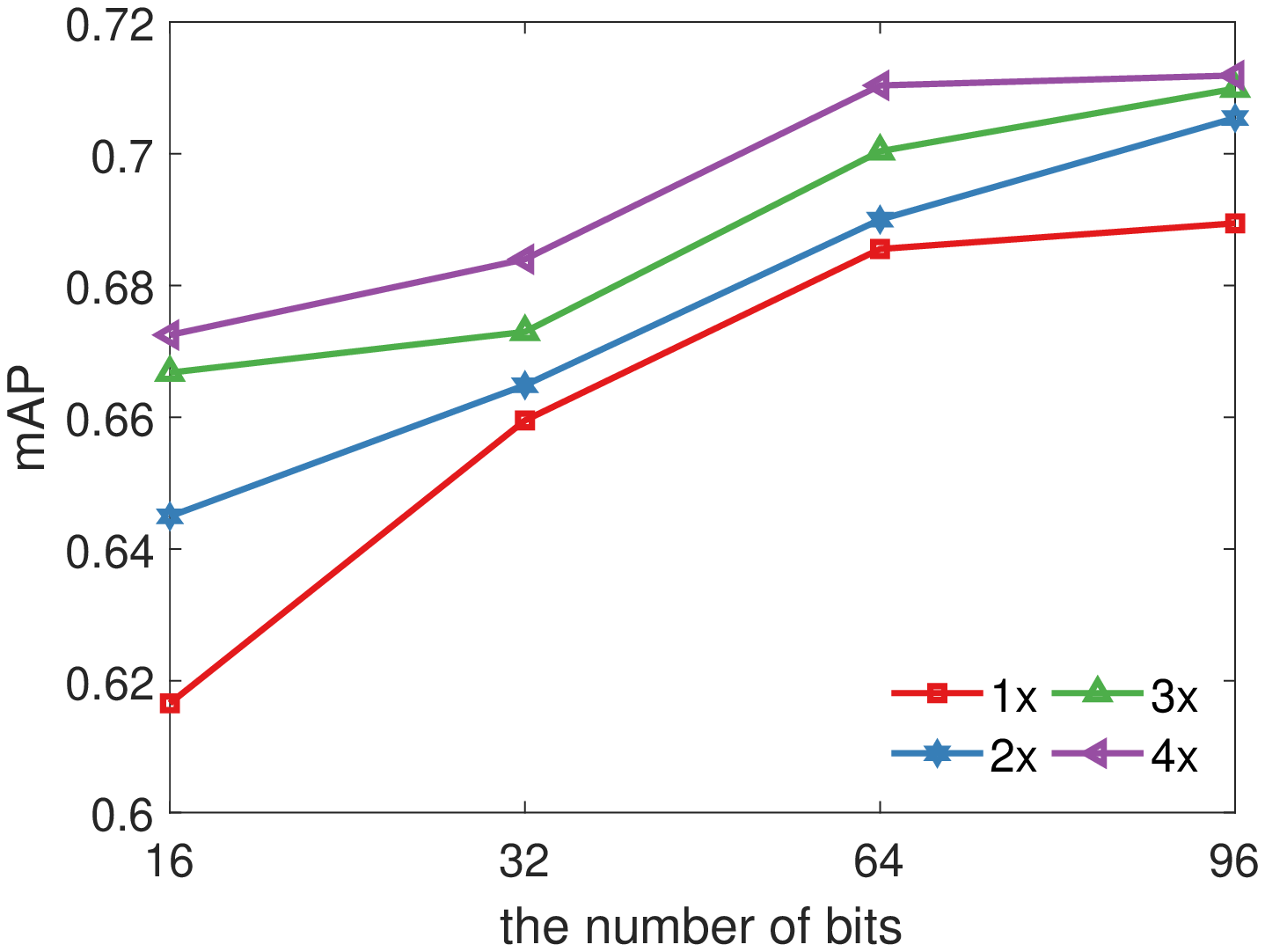}\\
{\small (a) MS-COCO}  &  {\small (b) NUS-WIDE}
\end{tabular}

\caption{The mAP results of different $l$.}
\label{fig:res7}
\end{figure}

By setting $C = 0.01$, we investigate the influence of the parameter $l$. Fig.~\ref{fig:res7} shows the comparison of OHSL by using different parameter values of $l$, where 1x denotes $l = b$ ($i.e.,$ the length of the binary target code is equal to the bit number), 2x denotes $l = 2b$, 3x denotes $l = 3b$ and 4x denotes $l = 4b$. With the increase of the length of the binary target code $l$, our method can achieve better accuracy. Although the accuracy is improved as the length of the binary target code increases, its time cost also rises. As the accuracy and the time cost are both important, to achieve a good balance between the accuracy and the time
cost, we take $l=3b$.

\begin{figure}[!htbp]
\centering
\begin{tabular}{cc}
 \includegraphics[width=0.46\columnwidth]{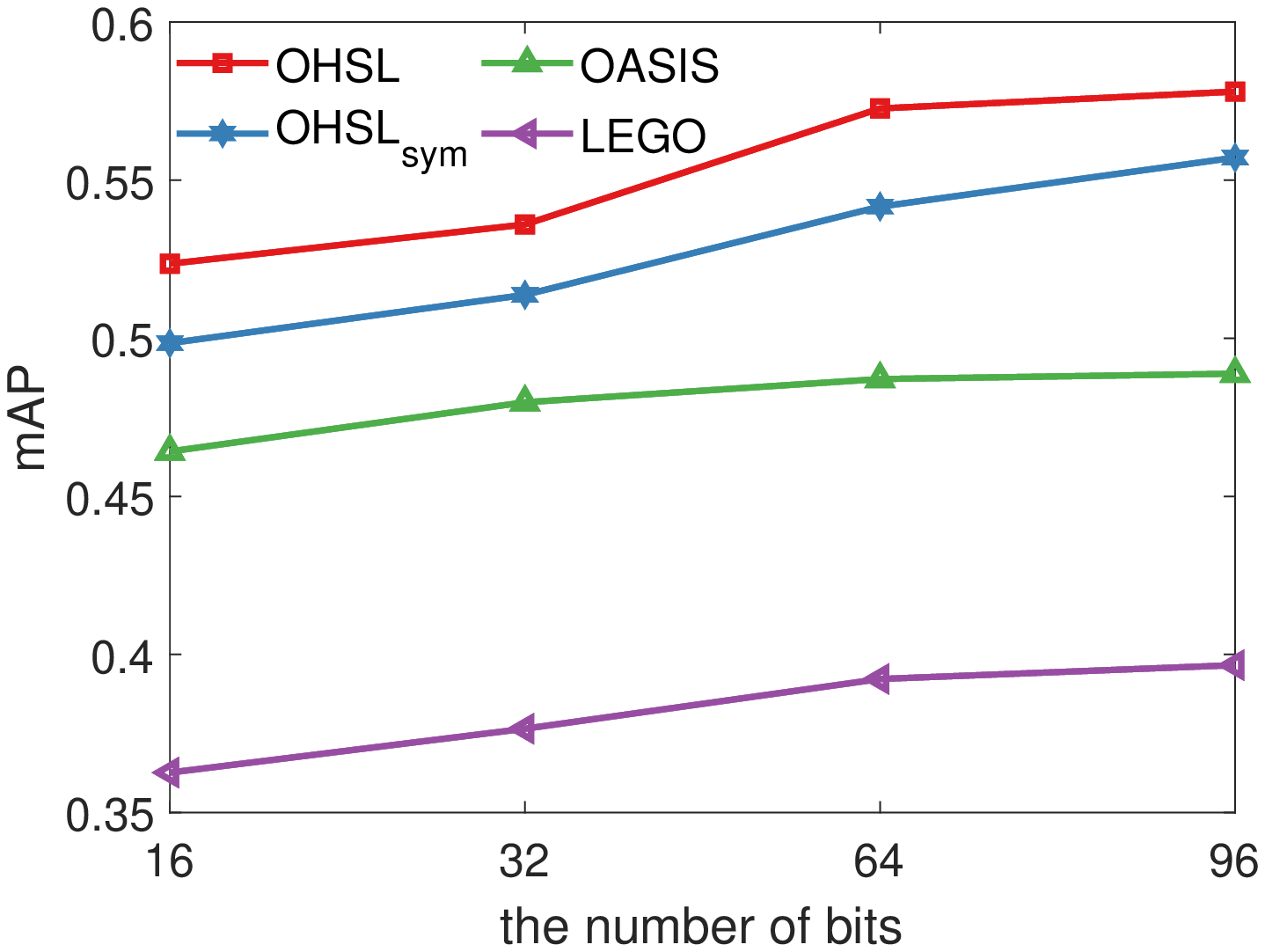}&
 \includegraphics[width=0.46\columnwidth]{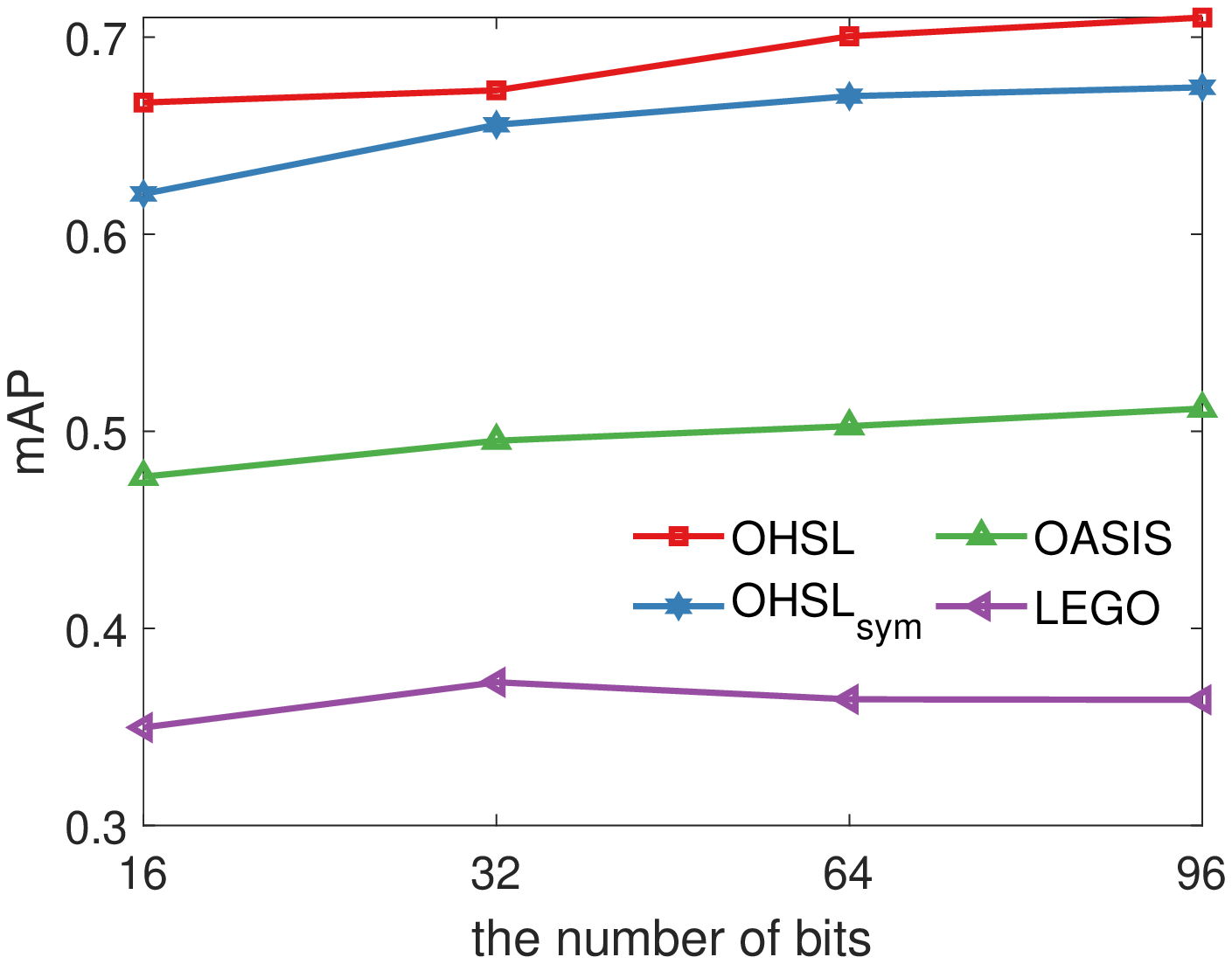}\\
{\small (a) MS-COCO}  &  {\small (b) NUS-WIDE}
\end{tabular}

\caption{The mAP results of different online learning ways.}
\label{fig:res8}
\end{figure}

Fig.~\ref{fig:res8} shows the comparison by using different ways to learn the bilinear similarity function. OHSL-sym denotes our method process the query and the binary codes symmetrically as shown in Eqn. (\ref{alg2}). OASIS~\cite{Chechik:2010:LSO:1756006.1756042} and LEGO~\cite{jain2009online} are the classical online metric learning methods to learn the square bilinear similarity by taking the binary codes as the input. By comparing OHSL to OHSL-sym, it shows that process the query and the binary codes asymmetrically can boost the search accuracy. And our method can achieve better accuracy than OASIS and LEGO.

\section{Conclusions}
In this paper, we propose a new online hashing framework without updating binary codes. By introducing the bilinear similarity function, the process of updating
binary codes is omitted and the similarity function is learnt online to adapt to the streaming data. A metric learning algorithm is proposed to learn the similarity function that can treat the query and the binary codes asymmetrically. The experiments on two multi-label datasets show that compared with the online hashing methods that need to update the hash functions, our method can achieve competitive or better search accuracy with much smaller time cost. Compared with the online hashing method that fixes the hash functions and learns the projection functions, our method outperforms it in terms of efficiency and accuracy.

\bibliographystyle{named}
\bibliography{egbib}

\end{document}